\documentclass[runningheads]{llncs}
\usepackage{subcaption}
\usepackage{booktabs}
\usepackage{graphicx}
\usepackage{amssymb}
\usepackage[colorlinks=true, allcolors=blue]{hyperref}
\usepackage{amsmath}
\usepackage[T1]{fontenc}
\usepackage{graphicx,verbatim}

\begin{document}

\title{Single Image Test-Time Adaptation via Multi-View Co-Training} 

\titlerunning{Single Image Test-Time Adaptation via Multi-View Co-Training}
\author{Smriti Joshi\inst{1} \and
Richard Osuala \inst{1} \and
Lidia Garrucho \inst{1}\and
Kaisar Kushibar \inst{1} \and
Dimitri Kessler \inst{1} \and 
Oliver Diaz \inst{1, 2} \and
Karim Lekadir\inst{1, 3}}

\authorrunning{S. Joshi et al.}

\institute{Departament de Matemàtiques i Informàtica, Universitat de Barcelona, Spain
\email{smriti.joshi@ub.edu}
\and
Computer Vision Center, Bellaterra, Spain
\and
Institució Catalana de Recerca i Estudis Avançats (ICREA), Passeig Lluís Companys 23, Barcelona, Spain
}

\maketitle            

\begin{abstract}
Test-time adaptation enables a trained model to adjust to a new domain during inference, making it particularly valuable in clinical settings where such on-the-fly adaptation is required. However, existing techniques depend on large target domain datasets, which are often impractical and unavailable in medical scenarios that demand per-patient, real-time inference. Moreover, current methods commonly focus on two-dimensional images, failing to leverage the volumetric richness of medical imaging data. Bridging this gap, we propose a Patch-Based Multi-View Co-Training method for Single Image Test-Time adaptation. Our method enforces feature and prediction consistency through uncertainty-guided self-training, enabling effective volumetric segmentation in the target domain with only a single test-time image. Validated on three publicly available breast magnetic resonance imaging datasets for tumor segmentation, our method achieves performance close to the upper bound supervised  benchmark while also outperforming all existing state-of-the-art methods, on average by a Dice Similarity Coefficient of 3.75\%. We will publicly share our accessible codebase, readily integrable with the popular nnUNet framework, at \url{https://github.com/smriti-joshi/muvi.git}. 
\end{abstract}


\section{Introduction} \label{sec:introduction}

 Domain shift in medical imaging presents a significant and well-known challenge that restricts the real-world applicability of automated methods 
due to the distribution gap between training and testing domains. This issue is attributed to differences in the acquisition hardware, field strength, reconstruction software, and imaging protocols. The differences may be exhibited in the overall quality and appearance, signal intensities and contrast, resolution and slice thickness, and field of view and acquisition plane of the images. While the clinicians are able to readily cope with domain shifts and account for them in their assessments, it is detrimental for deep learning based automatic methods. Typically, transfer learning \cite{kushibar2019supervised}\cite{zhang2020automated} and unsupervised domain adaptation methods \cite{joshi2021nn}\cite{dorent2023crossmoda} are used to alleviate the domain shift problem. However, both approaches rely on retraining with large target domain datasets at every domain shift occurrence. This requires knowledge of the specific type and magnitude of the shift, along with annotations that are consistent and free from inter-observer variability \cite{joshi2024leveraging}. For these reasons, test-time domain adaptation methods have gained momentum in recent years, as they enable the model to adapt dynamically to various types of unforeseen domain shifts during inference in an unsupervised manner \cite{liang2025comprehensive}.

The existing test-time adaptation methods can be roughly divided into the following four categories. \textit{Auxiliary-Task} based approaches, first introduced in Test-Time Training (TTT) \cite{ttt}, have a Y-shape model architecture for pretraining, with a shared feature extractor between the main task (e.g. classification, segmentation) and a self-supervised proxy task, i.e. rotation prediction. At test time, only the auxiliary task is trained to adjust the feature extractor to target domain \cite{ttt++,tttflow}. \textit{Normalization} based methods reduce dependence on source training by adapting the network through modifications to normalization layers. For instance, PTN \cite{ptn} uses the test batch statistics for batch normalization (BN) layers during inference. BNAdapt \cite{bnadapt} adjusts the source domain BN stats with test statistics. Tent \cite{tent} uses the current batch statistics and updates the affine parameters of BN layers through an entropy minimization objective. Next, \textit{Self-Training} based methods train during test-time via pseudolabels. For example, SHOT \cite{shot} adapts only the feature extractor through self-training while keeping the classifier frozen. MEMO \cite{memo} and AugMix \cite{augmix} use prediction consistency between augmentations. Finally, \textit{Prototype-based} methods leverage reducing distance to class prototypes \cite{t3a,shot} during test-time. To address the specific challenges of medical images, the existing methodologies have been adapted or new approaches have been proposed. 
For instance, FSEG \cite{fseg} adapts affine parameters of BN layers through contour regularization and nuclear norm losses. Working with single image setups, TTAS \cite{ttas} adapts affine parameters of BN layers by minimizing loss between shape priors from 2D slices to their 3D counterpart. 
Finally, InTent \cite{intent} combines the predictions obtained by sampling BN statistics between training and test statistics in an entropy-weighted fashion. 

This paper focuses on source-free single image test-time adaptation. Existing methods depend on shape priors and are typically validated on organ-specific tasks. Current approaches often require a large batch of data and fail to leverage three-dimensional (3D) data, despite working with volumetric modalities.  
Fang et al. \cite{multiview-2022} distill information between views to guide the learning process of each other for image synthesis. A similar approach pretrains swin transformers \cite{swinmm} through a cross attention block between different views. Inspired by these works, 
we introduce a novel method to adapt the network during test-time on a single image for shape-variable tumor segmentation task through \textit{Multi-View Co-Training}.  In summary, our contributions include: 
\begin{enumerate}
    \item Designing a novel source-free test-time domain adaptation method based on uncertainty-informed patch-based multi-view co-training.
    \item Conducting a comprehensive evaluation of our method on three heterogeneous, publicly available breast MRI datasets for tumor segmentation, where it surpasses previous state-of-the-art methods and achieves performance close to the supervised benchmark.
    \item Examining the impact of normalization layers on reducing generalization error in the single image test-time adaption setting. 
\end{enumerate}

\section{Methods} 

\subsection{Problem Definition}
\begin{figure*}
  \centering
  \includegraphics[width=\textwidth]{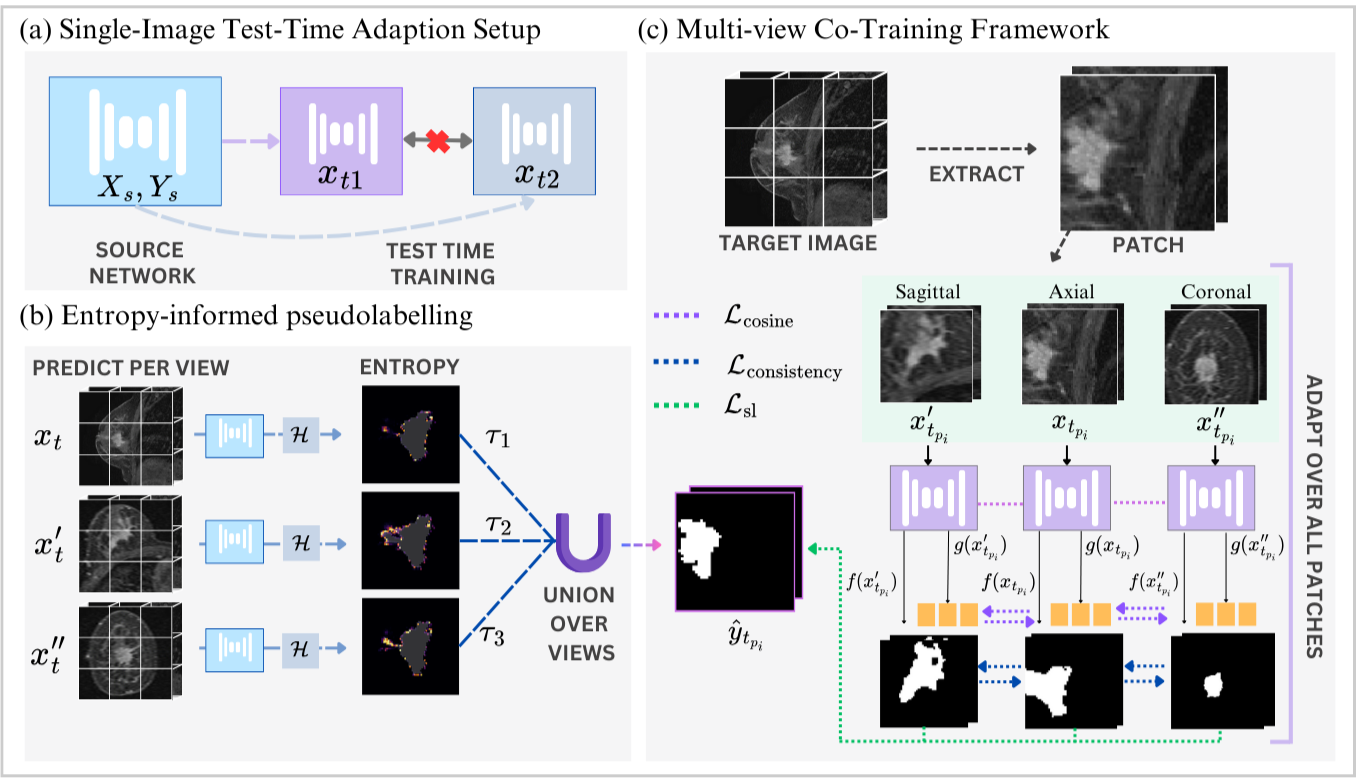}
  \caption {Pipeline of the proposed method MuVi. (a) Our setup where the source network is adapted independently for each target sample $x_{t}$, (b) computing pseudolabel through entropy-threshold union of prediction from each view, (c) Adaptation via patch-based training through our multi-view consistency framework.}
  \label{fig:graphical_abstract}
\end{figure*}

Given a model $M(\theta_{s})$  with parameters $\theta_{s}$   trained to learn function $f: X_{s} \rightarrow Y_{s}$  on source data $s$, our goal is to adapt $M(\theta_{s})$ to the target domain $t$ by learning the new function $k: X_{t} \rightarrow Y_{t}$, without access to the original source data 
nor the ground truth labels of the target samples. Here $X$ is a 3D tensor and the ground truth $Y$ is a densely-labeled voxel-wise 3D segmentation mask. Note that $M(\theta_{s})$ does not learn continually and is reset to $\theta_{s}$ for each $x_{t} \in X_{t}$, as is demonstrated in Figure \ref{fig:graphical_abstract}(a). Our decision to work on single images is motivated by the realistic scenario in hospitals where inference is needed on-demand and per-patient. Even if a delay in collecting the incoming batch is acceptable, a single target distribution cannot be guaranteed in the case of a single-center multi-scanner setup, rendering many existing techniques ineffective. Finally, medical imaging data consists of high-resolution volumes, requiring GPU processing with large VRAM capacities, making large batches expensive and resource-intensive.

\subsection{Test Time Adaptation}
Given target data $x_{t} \in X_{t}$ and model $M(\theta_{s})$ having learnt $f_{\theta_{s}}$, we use a combination of three distinct techniques to learn function $k_{\theta_{t}}$.

\textit{Modulation parameters and BN statistics:} All model parameters $\theta$ are adapted in a single epoch. We apply batch normalization in two steps: (i) normalization by current batch statistics, $\hat{x} = (x - \mu)/(\sqrt{\sigma^2 + \epsilon})$  where $\mu$ and $\sigma$ are mean and variance of the batch, and $\epsilon$ is a small constant for numerical stability and (ii) transformation by learnable parameters $\gamma$ and $\beta$ as $y = \gamma \hat{x} + \beta$. In multiple domain-shift scenarios such as changes in acquisition protocol or scanner types, BN statistics from the source network trained with diverse data can still provide a reasonable estimate of the target distribution as opposed to modifying them based on patches of a single image, which can adversely affect model performance. Therefore, we keep the original mean $\mu_s$  and variance $\sigma_s$ obtained during pre-training while adapting parameters $ \gamma $ and $\beta$ through gradient propagation of the loss function. 

\textit{Multi-view Co-Training:} Building upon intuition gathered from previous works \cite{multiview-2022,swinmm}, we propose patch-based multi-view training as a measure to adapt images during test time, thereby effectively mitigating the impact of diverse and a priori unknown domain shifts. To this end, for a test case $x_t$, we obtain the corresponding overlapping patches $\{ x_{t_{p_1}}, x_{t_{p_2}}, \dots, x_{t_{p_n}} \}$, where n is the total number of patches. Each patch is permuted to extract the corresponding representation from the remaining two views, $ x_{t_{p_i}}' = \pi_1( x_{t_{p_i}}), \quad  x_{t_{p_i}}'' = \pi_2( x_{t_{p_1}})$. The self-learning objective, $\mathcal{L}_{\text{sl}}$, is then minimized for each of these patches with pseudolabel $\hat{{y}}_{t{p_i}}$ (obtained as explained in the next section) as follows:

\[
\mathcal{L}_{\text{sl}} = \sum_{v \in \{ x_{t_{p_i}}, x_{t_{p_i}}', x_{t_{p_i}}'' \}} \left[\mathcal{L}_{\text{DICE}}(f(v), \hat{y}_{t_{p_i}}) + \mathcal{L}_{\text{CE}}(f(v), \hat{y}_{t_{p_i}}) \right],
\]

where DICE and CE stand for Dice Loss and Cross Entropy Loss, respectively. We further integrate a consistency constraint between views through  $\mathcal{L}_{\text{consistency}}$, by minimizing the same objective as above but between predictions from transformed views \( x_{t_{p_i}}' \), \( x_{t_{p_i}}'' \) and the original view \( x_{t_{p_i}} \). 
Additionally, we add a consistency term between feature embeddings of different views to ensure alignment independent of the pseudolabels. Specifically, we minimize
\[
\mathcal{L}_{\text{cosine}} = 1 - \cos(g(x_{t_{p_i}}), g(x_{t_{p_i}}')) + 1 - \cos(g(x_{t_{p_i}}), g(x_{t_{p_i}}'')),
\]
where \( g(\cdot) \) represents the feature extractor of the network. Consequently, the final loss is then given by \(\mathcal{L}_{\text{total}} = \lambda_1 \mathcal{L}_{\text{sl}} + \lambda_2 \mathcal{L}_{\text{consistency}} + \lambda_3 \mathcal{L}_{\text{cosine}}\), 

where we define \( \lambda_1, \lambda_2, \lambda_3 \) to calibrate the relative importance of each loss.  In our baseline implementation of this framework, we weigh all losses equally. 

\textit{Entropy-guided Self-Training:} To introduce complementary information from axial, coronal, and sagittal views, we compute an uncertainty-informed combination of corresponding predictions. This entropy-based pseudolabel is computed at the image level prior to training as opposed to the aforementioned patch-level multi-view co-training formulation. Specifically, the per-pixel entropy $H(z)$ is calculated using the predicted probabilities for each view, where $H(z) = -z \log_2(z) - (1 - z) \log_2(1 - z)$. Each prediction is subjected to an entropy threshold $\tau$, representing the minimum confidence required for accepting the prediction. These thresholds are empirically set on the validation, with values explored within the range of $[0.1, 0.6]$. A higher threshold ($0.4$) is applied to the view with the highest resolution, while a stricter threshold ($0.2$) is used for the other planes. The final entropy-based pseudolabel is given by the respective union as: 
\[
\hat{y} = \bigcup_{v \in \{ x_{t}, x_{t}', x_{t}'' \}} \{ j \mid H(\sigma(f(v(j))) < \tau_{v} \},
\]

where \( \sigma\) is the sigmoid function and $j$ refers to the image pixel. 
 

\section{Experiments and Results}

\subsection{Dataset and Implementation}

\textit{Dataset:} To train and validate our proposed method, we extract the first phase of the T1-weighted dynamic contrast-enhanced (DCE) sequence from three breast MRI datasets \cite{mama-mia}\cite{tcia}: 1) \textit{Duke-Breast-Cancer-MRI} \cite{duke_dataset}, collected between 2000-2014 and includes MRI scans of 922 patients acquired 
at a single center in the United States. 
    In this work, we only use the first phase of the DCE series from 254 cases for which the segmentation masks are available from \cite{mama-mia}. The DCE MRIs are acquired in the axial plane; 2) TCGA-BRCA \cite{tcga_dataset}, collected from 1999 - 2004, contains MRI scans of 80 patients from four different centers in the United States. The acquisition planes of the DCE MRIs are either sagittal or axial; 3) ISPY1 \cite{ispy1_dataset}, collected from 2002 - 2006, contains MRI scans of 161 patients from a single center in the United States, with DCE MRIs acquired in the sagittal plane. Additional dataset statistics are shown in Table \ref{tab:dataset-characteristics}. We adopt the Duke-Breast-Cancer-MRI as the source training dataset due to it having the highest number of cases, alongside varied protocols and scanners.

\begin{table}[]
\scriptsize
\caption{\textbf{Overview of datasets used in this work.} Each dataset, containing variable vendors, scanners, and protocols, is treated as an independent domain.}
\begin{tabular}{lcccccccc}
\toprule
Dataset                & Vendors              & Scanners & \multicolumn{3}{c}{Pixel Spacing} & \multicolumn{3}{c}{Slice Thickness} \\\midrule
                       &                      &          & mean      & std      & median     & mean      & std       & median      \\ \midrule
Duke-Breast-Cancer MRI & GE, Siemens          & 7        & 0.73      & 0.11     & 0.70       & 1.07      & 0.15      & 1.00        \\
ISPY1                  & GE, Siemens, Philips & 7        & 0.76      & 0.14     & 0.78       & 2.42      & 0.54      & 2.30        \\
TCGA-BRCA              & GE                   & 3        & 0.68      & 0.10     & 0.66       & 2.17      & 0.29      & 2.00      \\ \bottomrule 
\end{tabular}
\label{tab:dataset-characteristics}
\end{table}

\textit{Setup:} We train with unilateral breasts due to a lack of bilateral tumor annotations in the datasets. For each dataset, 30 cases are reserved for testing, and the remaining cases are used for model training with an 80:20 train-validation split. We use the 3D UNet 
from nnUNet \cite{nnunet} as the base framework for source training. This decision is motivated by the need to establish a strong baseline, utilizing the framework's robust pre-processing, prediction, and post-processing pipeline. The following modifications are made to the original pipeline: (i) the model is trained for 500 epochs, and (ii) the normalization layers are changed from \textit{Instance Normalization} to \textit{Batch Normalization}. 
The models are evaluated using three widely used segmentation metrics \cite{metrics_reloaded}: Dice Similarity Coefficient (DSC), Hausdorff Distance (HD) 
and Average Asymmetric Surface Distance (ASD).
All models are trained on a single NVIDIA GeForce RTX 4080 16 GB GPU.

\subsection{Results}
The results of the experiments on two datasets, TCGA-BRCA and ISPY1, are demonstrated in Tables \ref{tab:tcga} and \ref{tab:ispy1} respectively, where our method shows the largest improvement of $5.57\%$ and $4.13\%$ over the baseline in terms of DSC.\\

\textit{Baselines and Supervised Bounds:}  We compute two baselines, (i) anisotropic patch size of [112, 224, 96] (as computed by nnUNet preprocessing pipeline based on the input data characteristics) and (ii) isotropic patch size of [128, 128, 128]. By providing spatial consistency and avoiding directional biases through an isotropic input, nnUNet improves on the target domain test set by an average of 7.4\% DSC.
We use this baseline for performing test time adaptation with all methods. We also train supervised networks for computing two upper bounds, target only and source plus target, the latter achieving better performance, likely because of its larger training cohort. 
\begin{figure}
  \centering
  \includegraphics[width=\textwidth]{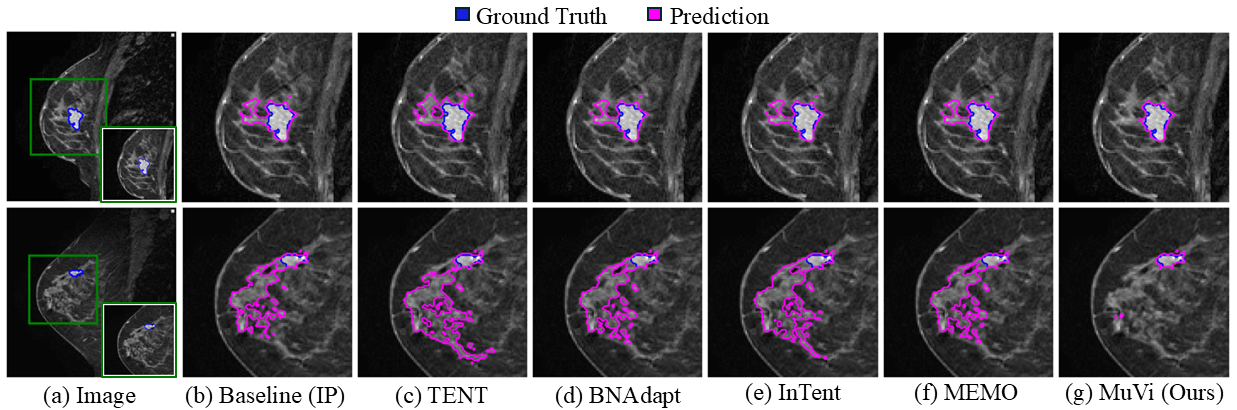}
  \caption {Qualitative segmentation results from different methods. Our method localizes the tumor precisely while removing the misidentified breast tissue.}
   \label{fig:qualitative_results}
\end{figure}

\begin{table}[h]
\scriptsize
\centering
\caption{Results of the mentioned methods on \textbf{TCGA-BRCA Dataset}. T refers to target only, S + T refers to source and target data combined. AP and IP refer to 3D UNet with anisotropic and isotropic patches. Best results in bold.}
\begin{tabular}{lccc}
\toprule
\textbf{Metric}                             & \textbf{DSC $\uparrow$}            & \textbf{HD$ \downarrow$}  & \textbf{ASD$ \downarrow$}                \\ \midrule
\textit{Upper Bound} \\
Supervised (T)                     & 0.6648 ± 0.2558         & 22.4165 ± 30.8837 &  16.1690  ± 26.1061     \\
Supervised (S + T)  & 0.7008 ± 0.2616         & 15.4179 ± 37.9761 & 11.6789 ± 29.1357        \\\midrule
\textit{Baseline} \\
3D UNet (AP)                       &0.5925 ± 0.3139          & 16.2708 ± 30.6949 &  14.4209 ± 29.4008 \\
3D UNet (IP)  & 0.6254 ± 0.269
          & 16.6877 ± 21.1614 & 12.5212 ± 16.7173
        \\ \midrule
\textit{Test-Time Adaptation}\\
PTN \cite{ptn}                              & 0.5187 ± 0.2764
         & 35.2485 ± 32.1232 & 26.5875 ± 24.8109
          \\
Tent \cite{tent}    & 0.5187 ± 0.2764
          & 35.2492 ± 32.1239 & 26.5879 ± 24.8103
          \\
BNAdapt \cite{bnadapt}     & 0.6418 ± 0.2509
          & 17.9102 ± 22.42 &  13.3762 ± 17.5268
          \\
InTent \cite{intent}   & 0.6228 ± 0.2689
          & 18.8721 ± 21.9771 &  14.3614 ± 17.3471
       \\
MEMO \cite{memo}                              & 0.6435 ± 0.2497
          & 17.787 ± 22.4028 & 13.2727 ± 17.5068        \\
MuVi (ours)               & \textbf{0.6811 ± 0.2269} & \textbf{15.9166 ± 25.6073} & \textbf{11.45655 ± 20.1114}
 \\ \midrule
\textit{Ablation: MuVi} \\
w/o source BN statistics             & 0.3813 ± 0.3262         & 57.1691 ± 48.2155
         & 52.2985 ± 47.3137 \\
w/o entropy-weighted labels          & 0.6341 ± 0.2652         & 22.7417 ± 32.4716
         & 17.9036  ± 27.7267\\
w/o consistency                      & 0.6745 ± 0.227          & 17.1451 ± 28.8524
         & 12.8935 ± 23.8826\\ \bottomrule
\end{tabular}
\label{tab:tcga}
\end{table}

\textit{Comparison with Other Methods:} We selected five state-of-the-art methods for detailed comparison on 
the clinically highly relevant tumor segmentation task. Auxiliary task-based methods (dependence on architecture) as well as prototype-based methods (dependence on shape) are excluded. We select the most popular state-of-the-art normalization-based methods, namely PTN \cite{ptn}, Tent \cite{tent}, BNAdapt \cite{bnadapt}, and InTent \cite{intent}. We also include the self-training based method MEMO \cite{memo}, which is closest to our approach since it adds consistency between augmentations during test time training. BNAdapt, InTent and MEMO are tested for single image adaptation in their original implementation and corresponding settings are used for comparison. Table \ref{tab:tcga} and \ref{tab:ispy1} show that our method outperforms existing techniques, demonstrating the effectiveness of our proposed entropy-based pseudolabels informed by three views and reduced reliance on BN layers. In Figure \ref{fig:qualitative_results}, we further provide a visual comparison, demonstrating how our method effectively distinguishes the tumor from surrounding breast tissue, accurately localizing its boundaries. 

\textit{Ablation of Entropy Weighted Pseudolabel and Consistency:} As shown in Table \ref{tab:tcga}, directly averaging predictions from all views without accounting for uncertainty still improves on the baseline. However, since the baseline is not trained on multiple views, selectively propagating only the confident labels from complementary lower-resolution views is essential for achieving substantial improvements. In our case, this approach leads to a 5.4\% increase in DSC and an approximately 7-point reduction in HD. Enforcing consistency across views enhances feature robustness to varying perspectives and pixel resolutions, resulting in further refinement of results in just one epoch. 

\begin{table}[]
\centering
\scriptsize
\caption{Results of the mentioned methods on \textbf{ISPY1 Dataset}. T Refers to target only, S + T refers to source and target data combined. AP and IP refer to 3D UNet with anisotropic and isotropic patches. Best results in bold.}
\begin{tabular}{lccc}
\toprule
\textbf{Metric}                  & \textbf{DSC $\uparrow$} & \textbf{HD $\downarrow$}  & \textbf{ASD $\downarrow$} \\\midrule
\textit{Upper Bound}\\
Supervised (T)                   & 0.7010 ± 0.2324       & 9.7768 ± 18.0119     &  8.1711 ± 15.7302 \\
Supervised (S + T)  & 0.7248 ± 0.2191      & 8.8684 ±  14.5787    &  6.3187 ± 9.8694\\ \midrule
\textit{Baseline}\\
3D UNet (AP)                      & 0.5426 ± 0.3136    & 28.1163 ± 58.3449     &  21.1092 ± 43.7914 \\
3D UNet (IP)                  & 0.6586 ±  0.2304       & 12.8004 ± 25.9020    &  10.5284 ± 24.7581 \\\midrule
\textit{Test-Time Adaptation} \\
PTN  \cite{ptn}                             & 0.6284 ± 0.2188        & 19.9790 ± 27.8741 & 14.3163 ± 19.3467     \\
Tent \cite{tent}                             & 0.6284 ± 0.2188        & 19.9792 ± 27.8760 & 14.3162 ± 19.3471   \\
BNAdapt \cite{bnadapt}                          & 0.6559 ± 0.2238        & 13.5064 ± 25.9386 &  10.9942 ± 24.9634   \\
InTent \cite{intent}                          & \textbf{0.6606 ± 0.2287}        & 10.0681 ± 14.7725 &  7.9732 ± 12.4688  \\
MEMO \cite{memo}                             & 0.6586 ± 0.2254        & 12.8885 ± 26.0073 &   10.6552 ± 25.0241   \\
MuVi (ours)                      & 0.6588 ± 0.2362       & \textbf{8.7271 ± 14.9046} &  \textbf{6.9601 ± 12.0732} \\\midrule
\textit{Experiment: Instance Normalization}\\
3D UNet (AP)                      & 0.5823 ± 0.2563        & 21.3682 ± 31.9623    & 16.1128 ±  23.8279 \\
3D UNet (IP)                   & 0.6564 ± 0.2625        & 9.3165 ± 16.4794      & 7.5948 ± 13.8091 \\ 
MEMO \cite{memo}                             & 0.6568 ± 0.2623        & 9.2749 ± 16.4329   & 7.5669 ± 13.7739  \\
MuVi (ours)                      &\textbf{ 0.7001 ± 0.2349}        & \textbf{6.1374 ± 11.2560} &  \textbf{5.1850 ± 9.7773} \\ \bottomrule
\end{tabular}
\label{tab:ispy1}
\end{table}

\textit{Effect of Normalization:} Emphasizing the importance of using source domain BN statistics during testing, we observe in metrics obtained from Tent \cite{tent} and PTN \cite{ptn} that relying on test image statistics worsens performance as a single image is insufficient for approximating the target domain distribution. InTent \cite{intent}, which is proposed to alleviate this very issue, also shows only a slight improvement in ISPY1 dataset while performing worse on TCGA-BRCA dataset. SAR \cite{sar} argues that BN layers can be ineffective in real-world scenarios and instead adopts Layer and Group Normalization. Taking a similar direction, we evaluate our method using Instance Normalization (IN) layers, a method well-suited for small batches and style variations such as in medical images. We compare our method to only MEMO \cite{memo} as other methods explicitly depend on the BN layers. 
As shown in Table \ref{tab:ispy1}, BN and IN baselines perform similarly on the test data. In contrast to MEMO which hardly improves on the baseline, our method, not bound by source BN stats, now achieves a notable 4.37\% increase in DSC and, surprisingly, HD and ASD that are even lower than the supervised benchmark.

\section{Discussion and Conclusion} \label{sec:discussion}
In this work, we propose a novel method for source-free test-time adaptation of 3D medical images. Through a multi-view co-training framework, we introduce complementary \textit{volumetric} information from axial, sagittal and coronal views by an entropy-informed self-training objective. Our method is label-free, adapts on a single image and needs only one epoch of training. We further demonstrate that a simple measure of using an isotropic patch size notably improves generalization by mitigating directional biases. We also investigate the role of batch normalization,  a key component in current literature that relies on large batches for adaptation, a requirement that is often difficult to meet in medical settings. Our findings suggest that batch-agnostic methods, i.e. Instance Normalization, offer greater stability in mitigating the effects of unseen domain shifts. We extensively validate our method on three publicly available multi-site multi-vendor breast-MRI datasets for tumor segmentation, demonstrating improvements over existing state-of-the-art methods.

\begin{credits}
\subsubsection{\ackname}
This project has received funding from European research and innovation programme under grant agreement No 101057699 (RadioVal) and Horizon 2020 under grant agreement No 952103 (EuCanImage). The work has also been supported by FUTURE-ES (PID2021-126724OB-I00) and AIMED (PID2023-146786OB-I00) from the Ministry of Science and Innovation of Spain.

\subsubsection{\discintname}
The authors have no competing interests to declare that are relevant to the content of this article.
\end{credits}

\bibliography{references}
\bibliographystyle{splncs04}

\end{document}